\documentclass{article}
\usepackage[left=0.75in,right=0.75in,top=1in,bottom=1in]{geometry}

\usepackage{amsmath,amsfonts}
\usepackage{algorithmic}
\usepackage{textcomp}
\usepackage{comment}
\usepackage{amssymb}
\usepackage{rotating}
\usepackage{pdflscape}
\usepackage{graphicx}
\usepackage{subcaption} 
\usepackage{enumitem}
\setlist[description]{style=nextline, font=\bfseries}
\usepackage{booktabs}
\usepackage{xcolor}
\usepackage{soul}
\usepackage[normalem]{ulem} 
\soulregister\sout7
\soulregister\ref7
\soulregister\cite7
\soulregister\gls7
\soulregister\etal7
\soulregister\eg7
\usepackage[framemethod=default]{mdframed}

\newcommand{\history}[1]{}
\newcommand{\doi}[1]{}

\newcommand{\received}[1]{}
\newcommand{\revised}[1]{}
\newcommand{\accepted}[1]{}
\newcommand{\published}[1]{}
\newcommand{\ieeecopyrighttext}[1]{}

\newcommand{\IEEEmembership}[1]{}
\newcommand{\IEEEauthorrefmark}[1]{}

\newcommand{\newtext}[1]{{#1}}
\newcommand{\oldtext}[1]{}

\newmdenv[
  linewidth=0.5pt,
  linecolor=blue!20,
  backgroundcolor=blue!5,
  roundcorner=3pt,
  innertopmargin=6pt,
  innerbottommargin=6pt,
  innerleftmargin=6pt,
  innerrightmargin=6pt
]{ieeebox}

\newcommand{\printRQa}{To what extent can general-purpose MLLMs 
perform binary iris PAD under \emph{minimal}, \ie naive prompting, relative to human examiners and a domain-specific state-of-the-art CNN baseline?}

\newcommand{\printRQaShort}{To what extent can general-purpose MLLMs 
perform binary iris PAD under naive prompting, relative to human examiners and a state-of-the-art CNN baseline?}

\newcommand{\printRQb}{By how much does a structured, task-scaffolded prompt improve the MLLMs' iris PAD performance?}

\newcommand{\printRQc}{Does injecting \emph{human salience} as in-context exemplars further improve MLLM's iris PAD performance as it does for CNN performance, and is the gain consistent across models?}

\newcommand{\printRQcShort}{Does injecting human salience improve MLLM's iris PAD performance, and is the gain consistent across models?}

\newcommand{\printRQd}{Can LLM-expanded human descriptions (introduced as MESH: Machine-Expanded Saliency from Human) produce a more complete salience signal that further enhances iris PAD performance and under which prompt regimes and base models does it help most?}

\newcommand{\printRQdShort}{Can LLM-expanded human descriptions (MESH: Machine-Expanded Saliency from Human)  further enhances iris PAD performance, and under which prompt regimes and base models does it help most?}

\newcommand{\etal}{\emph{et al.}\ }
\newcommand{\eg}{\emph{e.g.}, }

\newcommand{\ie}{\emph{i.e.}, }

\usepackage{bm}
\makeatletter
\AtBeginDocument{\DeclareMathVersion{bold}
\SetSymbolFont{operators}{bold}{T1}{times}{b}{n}
\SetMathAlphabet{\mathrm}{bold}{T1}{times}{b}{n}
\SetMathAlphabet{\mathit}{bold}{T1}{times}{b}{it}
\SetMathAlphabet{\mathbf}{bold}{T1}{times}{b}{n}
\SetMathAlphabet{\mathtt}{bold}{OT1}{pcr}{b}{n}
\SetSymbolFont{symbols}{bold}{OMS}{cmsy}{b}{n}
\renewcommand\boldmath{\@nomath\boldmath\mathversion{bold}}}
\makeatother

\usepackage[backend=bibtex]{biblatex}   
\begin{document}

\title{Generalist Multimodal LLMs Gain Biometric Expertise via Human Salience}

\author{Jacob Piland$^{1}$, Byron Dowling$^{1}$, Christopher Sweet$^{2}$, and Adam Czajka$^{1}$}

\date{}

\maketitle

\begin{center}
\small
$^{1}$Department of Computer Science and Engineering, University of Notre Dame, Notre Dame, IN 46556, USA\\
aczajka@nd.edu, jpiland@nd.edu, bdowlin2@nd.edu\\[0.4em]
$^{2}$Center for Research Computing, University of Notre Dame, Notre Dame, IN 46556, USA\\
csweet1@nd.edu\\[0.4em]
\textit{Corresponding author: Adam Czajka} — aczajka@nd.edu
\end{center}

\begin{abstract}
Iris presentation attack detection (PAD) is critical for secure biometric deployments, yet developing specialized models faces significant practical barriers: collecting data representing future unknown attacks is impossible, and collecting diverse-enough data, yet still limited in terms of its predictive power, is expensive. Additionally, sharing biometric data raises privacy concerns. Due to rapid emergence of new attack vectors demanding adaptable solutions, we thus investigate in this paper whether general-purpose multimodal large language models (MLLMs) can perform iris PAD when augmented with human expert knowledge, operating under strict privacy constraints that prohibit sending biometric data to public cloud MLLM services. Through analysis of vision encoder embeddings applied to our dataset, we demonstrate that pre-trained vision transformers in MLLMs inherently cluster many iris attack types despite never being explicitly trained for this task. However, where clustering shows overlap between attack classes, we find that structured prompts incorporating human salience (verbal descriptions from subjects identifying attack indicators) enable these models to resolve ambiguities. Testing on an IRB-restricted dataset of 224 iris images spanning seven attack types, using only university-approved services (Gemini 2.5 Pro) or locally-hosted models (e.g., Llama 3.2-Vision), we show that Gemini with expert-informed prompts outperforms both a specialized convolutional neural networks (CNN)-based baseline and human examiners, while the locally-deployable Llama achieves near-human performance. Our results establish that MLLMs deployable within institutional privacy constraints offer a viable path for iris PAD.
\end{abstract}


\bigskip

\noindent\textit{This material is based upon work supported by the U.S. National Science Foundation under grant No. 2237880 and work supported by the U.S. Department of Defense under contract No. W52P1J-20-9-3009. Any opinions, findings, conclusions, or recommendations expressed in this material are those of the authors and do not necessarily reflect the views of the U.S. National Science Foundation or U.S. Department of Defense. The U.S. Government is authorized to reproduce and distribute reprints for Government purposes, notwithstanding any copyright notation hereon.}

\begin{figure*}[!ht]
    \centering
    \includegraphics[width=\linewidth]{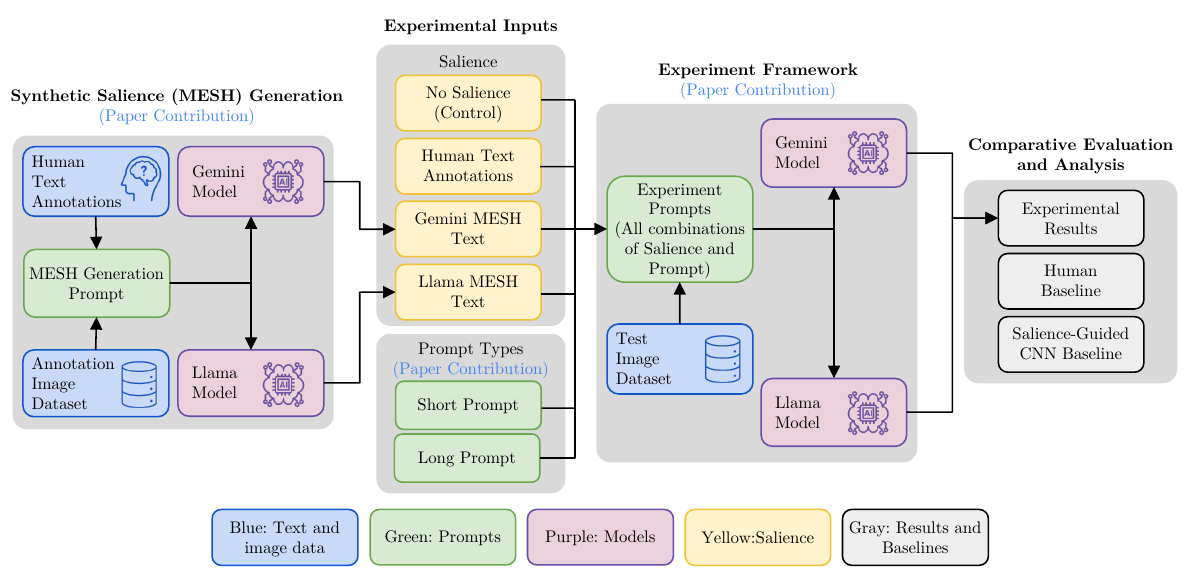}
    \caption{Experimental pipeline and paper contributions. We start with the generation of synthetic MESH salience from human annotations and corresponding image dataset. This salience, along with the control and human salience, is then combined with our novel prompts for testing. Results are obtained for both models to which we have ethical access and compared with a human and salience-guided CNN baseline.}
    \label{fig:teaser}
\end{figure*}

\section{Introduction}
\subsection{Background and Motivation}
Biometric presentation attack detection (PAD) stands at a critical intersection of security necessity and practical constraints. While deep learning approaches have shown promise for detecting various attack types ranging from artificial objects presented to biometric cameras to synthetic imagery, their deployment faces four fundamental challenges that motivate our investigation into alternative approaches, leveraging recent advances in foundational and multi-modal large models.

First, the diversity of presentation attacks requires extensive data collection across multiple attack vectors, which is expensive, ethically complex when involving human subjects, and will never exhaust all possible attack types potentially to be seen in the future. \oldtext{Second, the rapid evolution of attack methodologies, particularly with advances in generative AI, demands systems that can adapt quickly, without complete retraining.} \newtext{Second, the rapid evolution of attack methodologies, particularly with advances in generative AI, demands systems that can adapt quickly, without complete retraining \cite{yadav2024synthesizing}.} Thus, an effective PAD solution should ideally be based on models with a general ``understanding'' of visual tasks, and that are only directed by limited data (multi-modal, if available) towards the PAD task. \oldtext{Large multi-modal (vision-language) models may offer such capability.} \newtext{Large multi-modal (vision-language) models may offer such capability \cite{shahreza2025foundation}.} Third, while biometric PAD algorithms in general surpass human-sourced classification, there is an important application of forensic expert-based judgment in court cases, where the machine serves an important supporting role, but the final decision is left -- by law -- for the expert. 
In such cases there is a need for seamless human-machine pairing approaches. Fourth, the sensitive nature of biometric data creates significant barriers to model development and deployment. Biometric images are personally identifiable information subject to stringent privacy regulations and Institutional Review Board (IRB) restrictions that explicitly prohibit transmission to public cloud services such as ChatGPT or Claude. These privacy-related constraints eliminate many state-of-the-art/commercial solutions from consideration and there's a need for privacy-safe, usually locally-run approaches.


\subsection{Proposed Solution}
\oldtext{The standard approach of training specialized CNNs on large} \oldtext{datasets became impractical, and inference using modern} \oldtext{cloud-based AI tools may be legally prohibited if we process} \oldtext{biometric data. We thus propose a fundamentally different} \oldtext{approach:}\newtext{Due to the impractically of attaining ever larger biometric datasets, which is the standard approach to training specialized CNNs, we propose a fundamentally different approach:} leveraging the latent visual understanding of general-purpose \oldtext{multi-modal large language models (MLLMs)} \newtext{MLLMs} that can be deployed within institutional privacy boundaries. Our investigation is constrained to models that either have special institutional agreements for secure processing (Gemini 2.5 Pro through our university's bilateral agreement with Google) or can be hosted entirely on local infrastructure (e.g., Llama 3.2-Vision). This constraint, while limiting our model choices, reflects real-world deployment scenarios where biometric data cannot leave institutional control. In this work, we demonstrate the solution for iris PAD as the example PAD domain, although the proposed methodology can be applied to design a PAD method for other biometric techniques, for which both images and language descriptions of these images are available.

\begin{figure*}[!ht]
    \centering
    \subfloat{
        \includegraphics[width=0.48\textwidth]{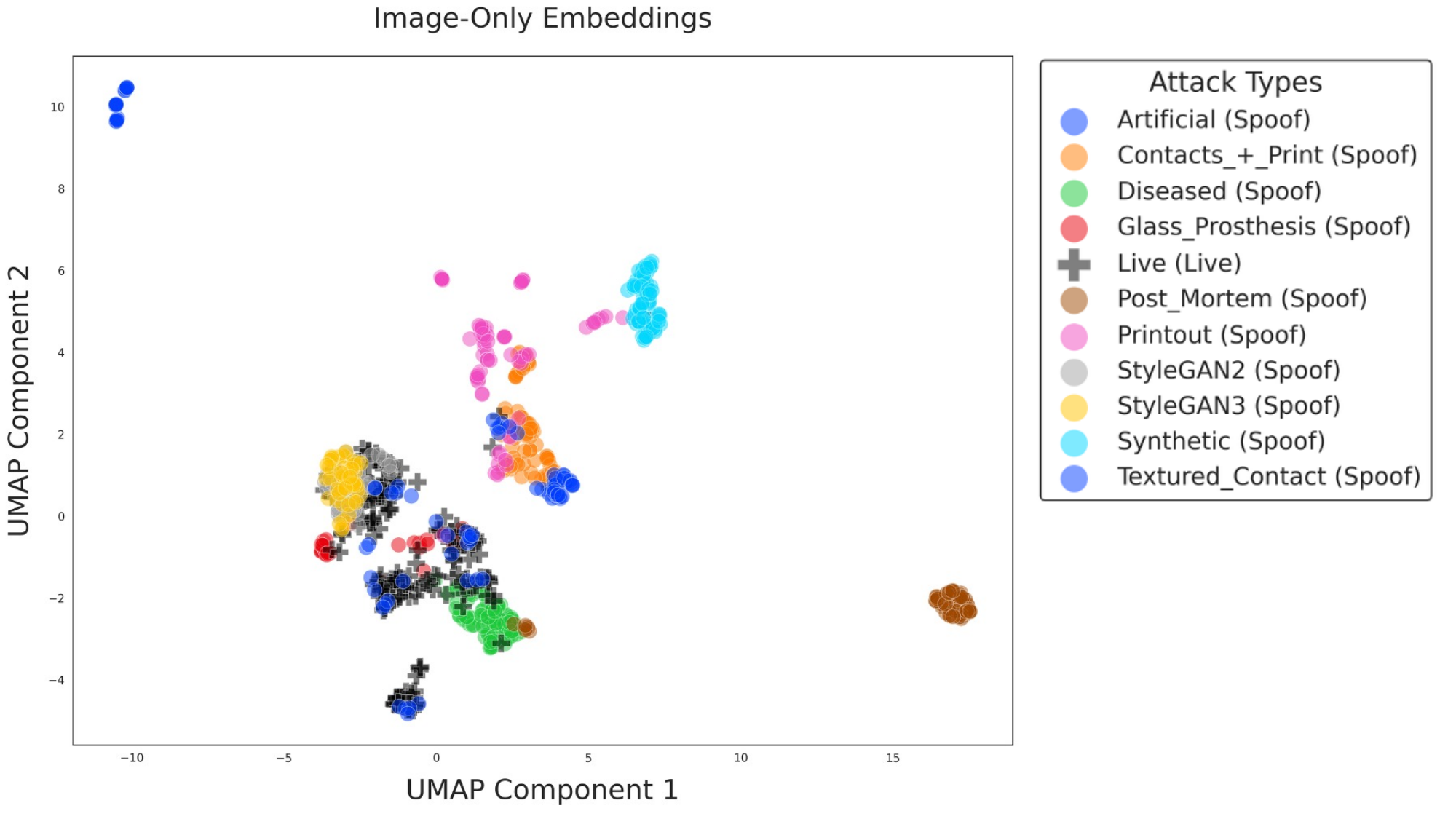}
    }
    \hfill
    \subfloat{
        \includegraphics[width=0.48\textwidth]{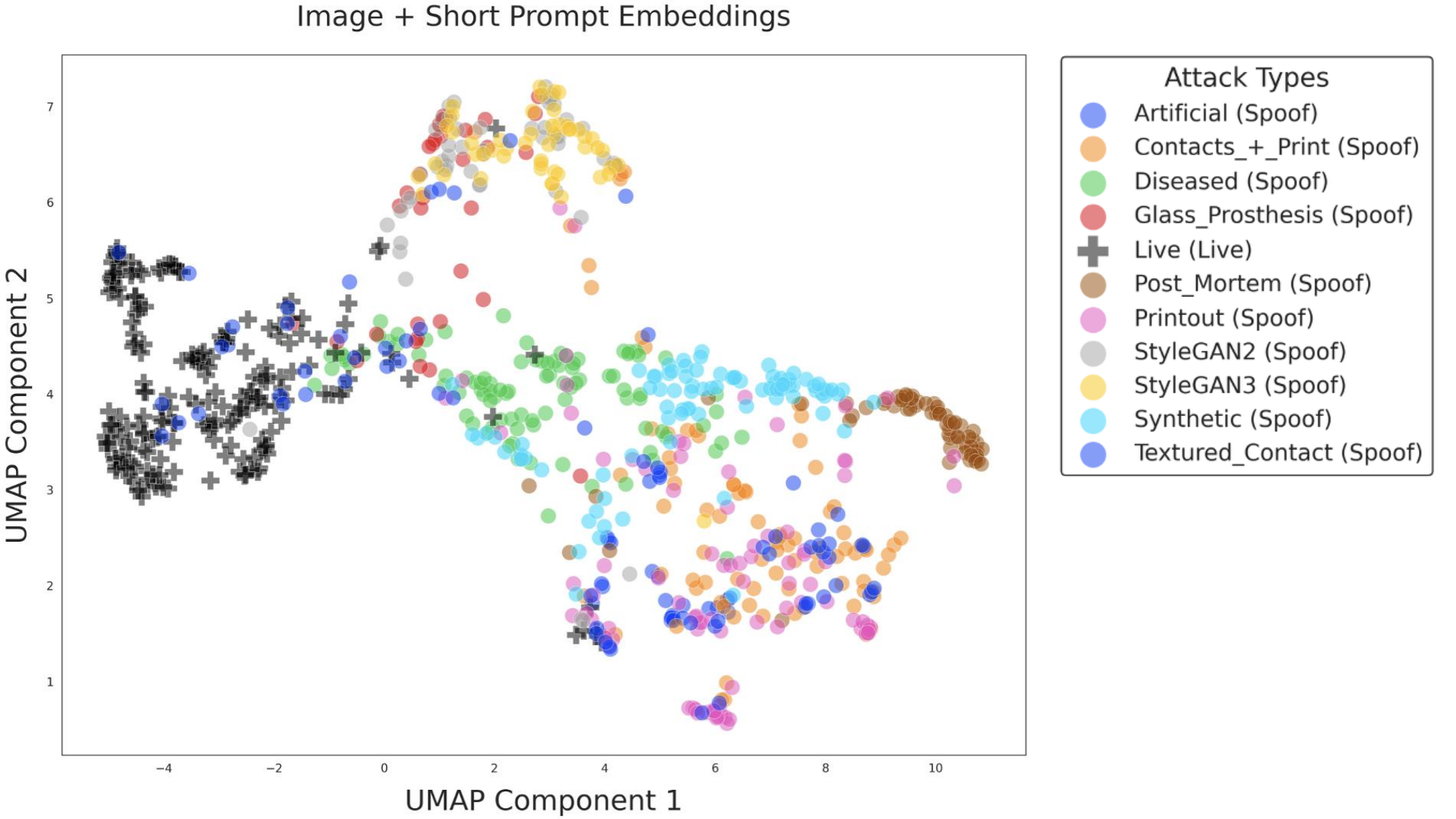}
    }
    \caption{Uniform Manifold Approximation and Projection (UMAP) visualization of iris samples encoded by (left) SigLIP vision-only embeddings and (right) SigLIP + Gemma multimodal embeddings using a simple binary prompt asking whether the iris is ``real and healthy'' or ``synthetic/unhealthy.'' Despite never being trained specifically for iris PAD, SigLIP alone achieves partial separation of attack types. However, adding even minimal semantic guidance through Gemma shows much improved visual separation between live vs. spoof discrimination, with clearer cluster boundaries and reduced overlap between classes. This visual separation motivates our investigation into whether general-purpose MLLMs can address specialized biometric security tasks through appropriate prompting.}
    \label{fig:umap_image_short}
\end{figure*}

Our key insight is that privacy-compliant MLLMs, despite never being explicitly trained on iris PAD tasks, have learned visual features that partially separate different attack types. We illustrate this through exploratory Uniform Manifold Approximation and Projection (UMAP) visualization \cite{mcinnes2018umap} using SigLIP \cite{zhai2023sigmoid}, a widely-used vision-language model, for image encoding and Gemma \cite{gemma2024}, a large language model, for text embedding. To leverage both SigLIP's strong visual representations and Gemma's text processing capabilities (as SigLIP is limited to short prompts), we combine their embeddings through a shallow \newtext{Multi-Layer Perceptron (MLP)} fusion layer, a standard approach in multimodal learning \cite{baltrusaitis2019multimodal}. \newtext{The shallow MLP is a two-layer network with normalization and GELU activation, an input shape of 2048 and a hidden dimension size of 512.} This combined representation, shown in Fig. \ref{fig:umap_image_short}, uses only a simple binary prompt asking whether the iris is 'real and healthy' or 'synthetic/unhealthy'\footnote{Full prompt: 'Is this a real and healthy iris or synthetic/unhealthy iris? Return a single float number from 0 to 1 with 0 being real/healthy and 1 being synthetic/unhealthy and no other output.' \newtext{The (un)healthy specification is in reference to the diseased iris attack type, which the model will not detect otherwise as diseased irises are real irises in a layman's context.}}, yet provides strong clustering of iris attack types.
While this natural clustering provides a strong foundation, we observe overlapping regions where attack types are not clearly distinguished by visual features alone. This observation suggests that the challenge extends beyond visual discrimination to include domain-specific interpretation of ambiguous cases.

To address this challenge, we introduce a framework that incorporates human salience, specifically verbal descriptions of attack indicators from both expert and non-expert observers, directly into MLLM prompts. This approach transforms the PAD task from one requiring extensive training data to one of knowledge transfer, where human expertise guides pre-trained models toward domain-specific decisions, particularly in cases where visual features alone are insufficient.

Working within strict IRB constraints that limited us to 224 images across seven attack types and restricted us to institution-approved or locally-hosted models, we demonstrate that this approach not only works but can exceed both specialized CNN baselines and human expert performance. At the same time, this work does not argue for replacing specialized PAD systems in high-throughput production environments. Instead, we establish feasibility for scenarios where traditional approaches face insurmountable barriers, particularly where privacy regulations prohibit cloud-based processing, offering a new tool for the biometric security community when conventional methods cannot be applied.

\subsection{Research Questions}

Our investigation and presented results are organized around the following four research questions:

\begin{description}[leftmargin=1cm]

\item[{\bf RQ1:}] \printRQa


\item[{\bf RQ2:}] \printRQb


\item[{\bf RQ3:}] \printRQc


\item[{\bf RQ4:}] \printRQd


\end{description}

\subsection{\newtext{Biometric Data and Public MLLM Services}}

\newtext{Due to the large and growing number of MLLM public services and their impressive generalization capabilities, there is an increasing body of work in which biometric data is submitted to such services. However, since many research biometric datasets were collected before the MLLM era, the consent forms used during those collections may not contain appropriate provisions to protect subjects’ privacy in the context of large commercial models. This ethical and privacy aspect of biometric data usage in the modern AI environment warrants short commentary.

As an example, the consent forms used by the University of Notre Dame, whose data was used in these experiments, contain several provisions to protect subjects’ privacy. These provisions forbid licensees from redistributing the data to third parties, using the data commercially, or using the data in a way that could cause subject embarrassment or mental anguish. The data-sharing license, which must be duly executed by the institution requesting a copy of the data, also requires licensees to obtain the Principal Investigators’ permission to publish more than ten images from the dataset in a paper.

None of these provisions can be guaranteed when data is submitted to public MLLM services without having the executed data sharing agreement executed by the MLLM service provider, which makes fulfilling all provisions included in the consent form impossible. In the case of the authors' institution, such an agreement was secured with Google Inc. for the use of the ``Gemini'' service. Consequently, this work reports only on results obtained with either locally-maintained open-source MLLMs or the commercial model available via the ``Gemini'' service.}

\section{Related Work}

\subsection{Detection of Biometric Presentation Attacks}

Presentation attacks are attempts at manipulating a biometric system into an incorrect decision in one of two forms: impersonation, where the objective is to deceive the biometric system with a positive match, or identity concealment, where the attacker wishes to evade detection entirely \cite{czajka2018presentation}. Examples of iris presentation attacks include synthetic iris imagery using Generative Adversarial Networks \cite{karras2020analyzing, karras2021alias, tinsley2023iris}, presenting a cadaver post-mortem sample to the iris scanner \cite{trokielewicz2018presentation}, printing a high-resolution image of an authentic iris and presenting it to the scanner \cite{Czajka_MMAR_2013}, or presenting an artificial eye such as a glass prosthesis \cite{czajka2018presentation}. 

Many current PAD techniques rely on deep-learning based approaches
\cite{boyd2022human, boyd2023cyborg, Nguyen_CSUR_2024, crum2025mentor}. \newtext{More recent works leverage on generalization capabilities of image foundation models \cite{Tapia_FG_2025}, or proposed new loss functions for image-text multimodal alignment \cite{Zhang_ASIG_2024} in the context of iris PAD.} What is currently underexplored and covered in the next section, is leveraging the powerful capabilities of \newtext{large multimodal} foundational models, which -- by seeing vast amount of data during training -- acquire capabilities of solving general vision tasks, and may offer an important component of few-shot learning approaches adapting such models to domain-specific tasks. This paper specifically explores application of multimodal large language models towards the task of iris PAD.

\subsection{Foundation Models and Iris Biometrics}

\oldtext{Multimodal Large Language Models (MLLMs)} \newtext{MLLMs} have many definitions but the consensus is that \oldtext{MLLMs} \newtext{they} are multimodal AI models that are capable of interpreting image and text typically by tethering a vision encoder to a Large Language Model (LLM) \cite{bordes2024introduction}. As newer versions of MLLMs have been released with increasingly larger parameter counts, researchers have attempted to push their capabilities into the realm of biometrics with promising results \cite{farmanifard2024chatgpt, sony2025benchmarking, hassanpour2024chatgpt, deandres2024good}. Farmanifard \etal explored GPT-4's abilities on zero-shot evaluation of iris matching tasks, cross-modality matching, and PAD finding that GPT-4 often succeeded in distinguishing iris pairs under a variety of challenging conditions such as occlusion and noisy environments \cite{farmanifard2024chatgpt}. 

Sony \etal evaluated 41 MLLMs on biometric tasks, including limited PAD, hypothesizing that the zero-shot baseline performance of these models can be improved with the addition of a classification head. Using a dataset consisting of both full size and cropped images of bonafide live irises and irises wearing patterned contact lenses, the authors measured the baseline zero-shot inference accuracy for foundation models. Next testing three types of classifiers, support vector machines with linear and RBF kernels as well as logistic regression, most models when trained on these embeddings showed a moderate to substantial jump in classification accuracy with DINOv2 \newtext{\cite{oquab2024dinov2}} and DINO-ViTB16 \newtext{\cite{caron2021}} showing the best performance on full size and cropped irises, respectively. 

Despite the success of Farmanifard's and Sony's works demonstrating the capabilities of MLLMs for PAD, the authors only tested two attack types, and for the purposes of their studies avoided directly prompting the model with the PAD task. Therefore it would be an interesting follow-up to test the capabilities of these MLLMs on a wider variety of attack types and investigate the effect of different prompt techniques and whether the injection of human salience would have an effect on model performance, which is the topic of this paper.

\subsection{Prompt Engineering}
\textit{Zero-shot prompting} is the core building block, from which other prompt engineering techniques derive as it is the core functionality of foundational models, \ie, directly asking the model to perform a task or query without any prior examples or injection of knowledge. Closely related to this are the concepts of \textit{few-shot prompting} and \textit{many-shot prompting} which are both forms of in-context learning where the model is fed high-quality input-output examples to better understand a task \cite{sahoo2024systematic, agarwal2024many}. Brown \etal found that on a variety of natural language processing (NLP) benchmarking datasets, few-shot learning often beats zero-shot learning by a significant margin and in some cases even beats SotA fine-tuned models such as on the LAMBADA dataset \cite{brown2020language, paperno2016lambada}.

\textit{Chain-of-Thought} (CoT) is a form of instructional prompting that guides the AI model through a series of intermediate reasoning steps to elicit a better response and significantly improve the model's responses on complex queries \cite{wei2022chain}. In zero-shot CoT prompting, this can be as simple as appending the phrase ``Let's think step by step'' to the end of a logic puzzle or a math problem, while few-shot CoT prompting involves providing examples of how to approach said problems step-by-step using this CoT reasoning. Wei \etal tested few-shot CoT prompting against the standard few-shot prompting popularized by Brown \etal \cite{brown2020language} by comparing the techniques on multiple benchmarks across five different LLM models. Their results indicate that CoT prompting outperforms few-shot prompting and in some cases beat SotA fine-tuned models, particularly on multi-step reasoning tasks, on larger models.

Finally \textit{Retrieval Augmented Generation} (RAG) is a technique that enhances a model's output by retrieving relevant information from an external knowledge base and injecting this knowledge into the prompt prior to generating a response from the model \cite{lewis2020retrieval}. The LLMs' ``memory'' is affected by a fixed cut-off data at the time of training and they can be prone to confabulations, \ie, generating plausible but factually inaccurate information \cite{wei2022emergent,huang2025survey}. RAG offers a method to improve upon these issues by continuously maintaining a specialized knowledge base to enhance the LLM's baseline capabilities without the need for retraining or fine-tuning the entire model \cite{lewis2020retrieval}.

\subsection{Human Salience}

Previously, the inexplicability of deep learning methods, in contrast to LLMs, has been alleviated by comparison to observed, human experts \cite{42richardwebster2018visual}.
In many tasks, machine learning accuracy is essentially always at least as good as human accuracy \cite{37o2012comparing}, however, human psychophysics has aided in deep learning tasks such as handwriting \cite{18grieggs2021measuring}, natural language processing \cite{53zhang2020human}, and scene description \cite{20he2019human,23huang2021specific}.
Specifically in biometrics (including iris PAD), human saliency has been shown to compliment machine saliency \cite{boyd2023cyborg, boyd2022human}. 

\begin{figure*}[!htb]

    \centering
    \begin{minipage}{0.23\textwidth}
        \includegraphics[width=\linewidth]{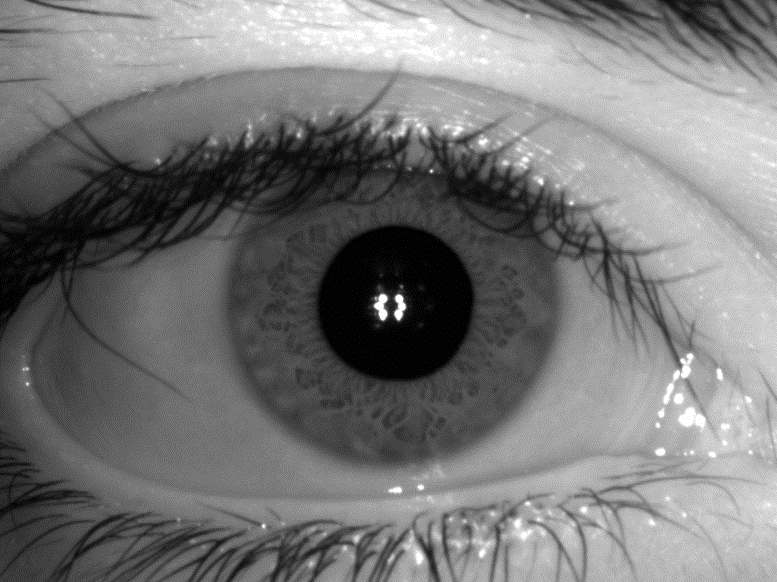}
    \end{minipage} \hspace{0.01\textwidth}
    \begin{minipage}{0.23\textwidth}
        \includegraphics[width=\linewidth]{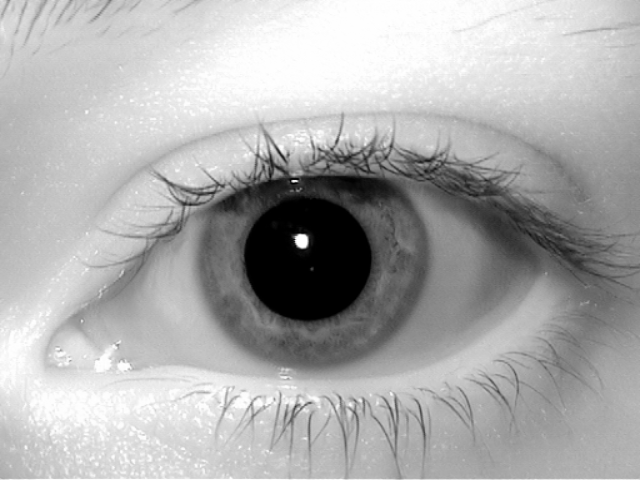}
    \end{minipage} \hspace{0.01\textwidth}
    \begin{minipage}{0.23\textwidth}
        \includegraphics[width=\linewidth]{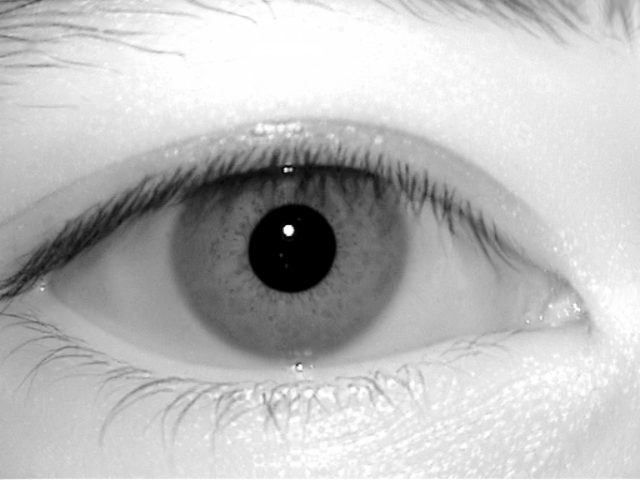}
    \end{minipage} \hspace{0.01\textwidth}
    \begin{minipage}{0.23\textwidth}
        \includegraphics[width=\linewidth]{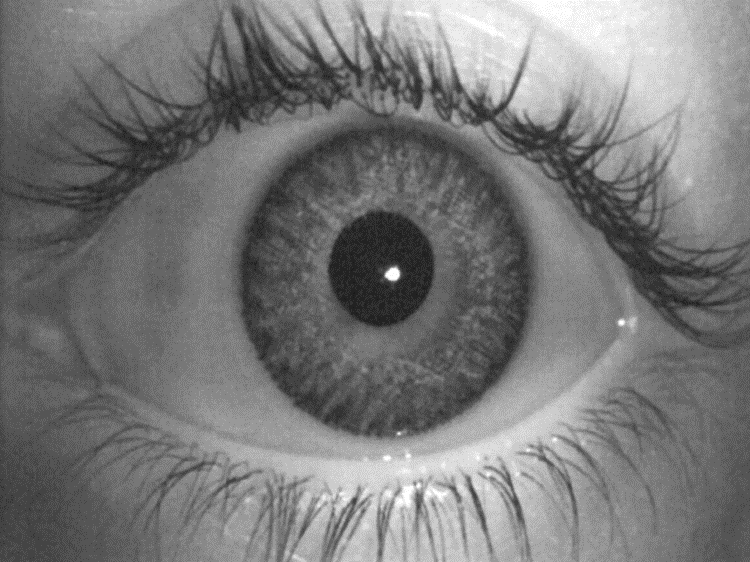}
    \end{minipage}

    \vspace{0.2cm}

    \begin{minipage}{0.23\textwidth}
        \includegraphics[width=\linewidth]{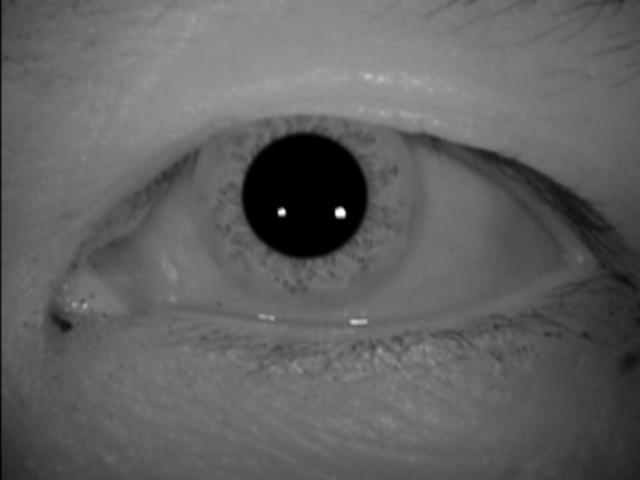}
    \end{minipage} \hspace{0.01\textwidth}
    \begin{minipage}{0.23\textwidth}
        \includegraphics[width=\linewidth]{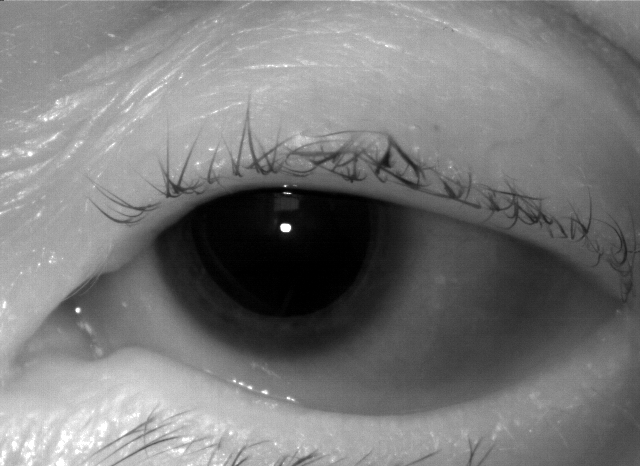}
    \end{minipage} \hspace{0.01\textwidth}
    \begin{minipage}{0.23\textwidth}
        \includegraphics[width=\linewidth]{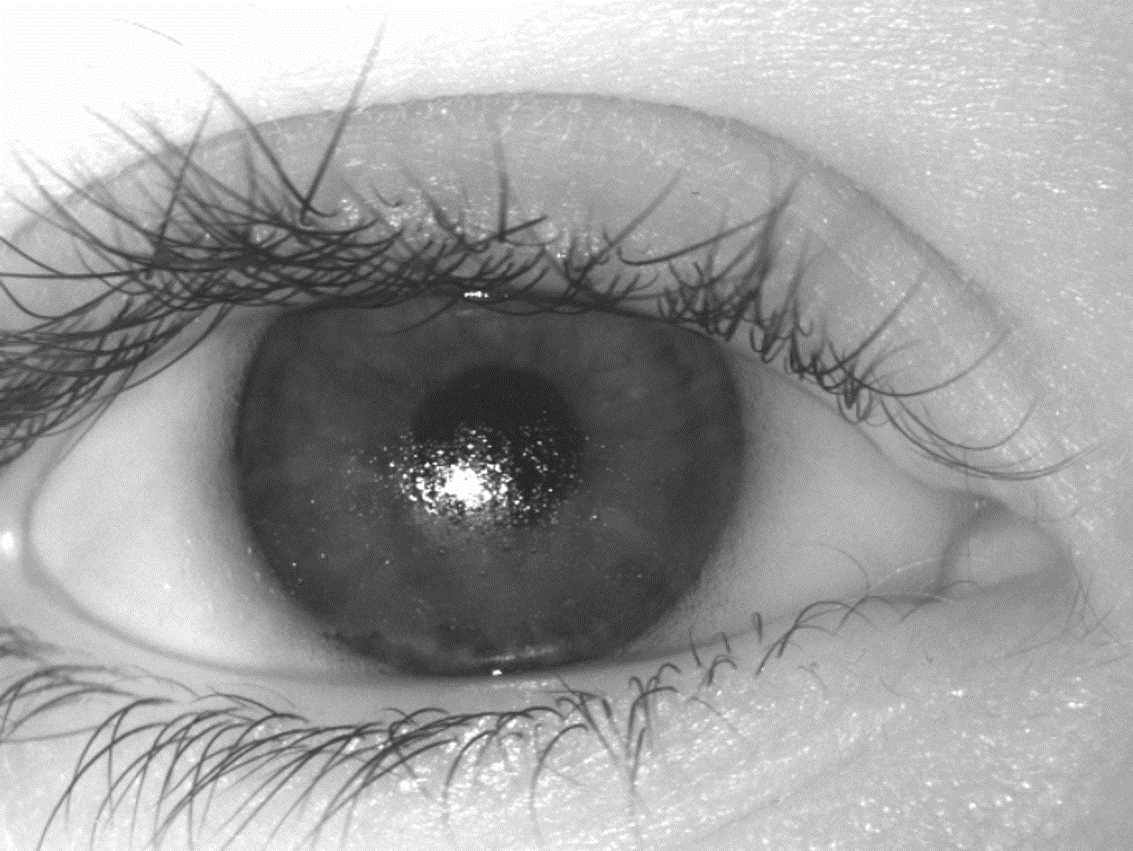}
    \end{minipage} \hspace{0.01\textwidth}
    \begin{minipage}{0.23\textwidth}
        \includegraphics[width=\linewidth]{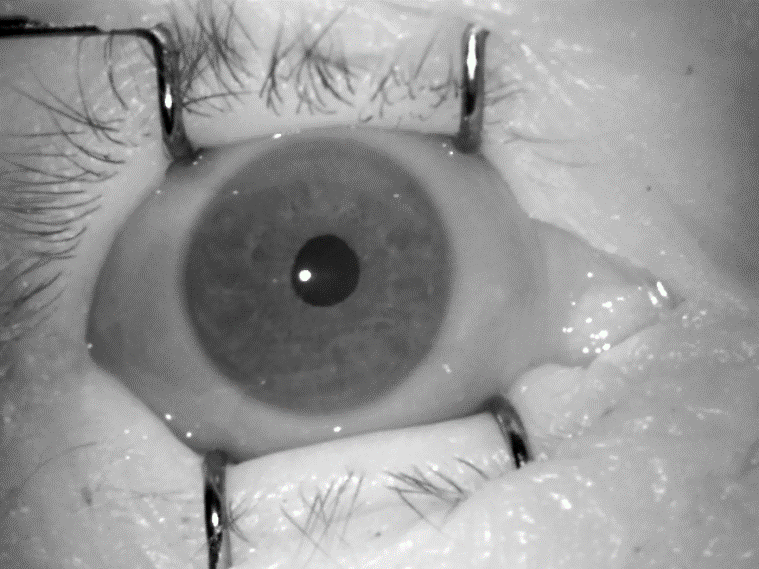}
    \end{minipage}

    \vspace{0.2cm} 

    \begin{minipage}{0.23\textwidth}
        \includegraphics[width=\linewidth]{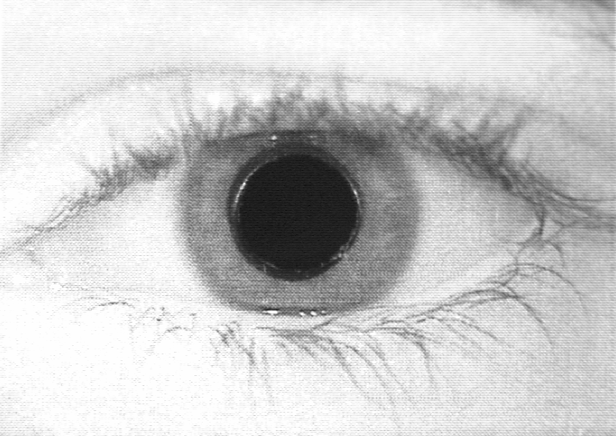}
    \end{minipage} \hspace{0.01\textwidth}
    \begin{minipage}{0.23\textwidth}
        \includegraphics[width=\linewidth]{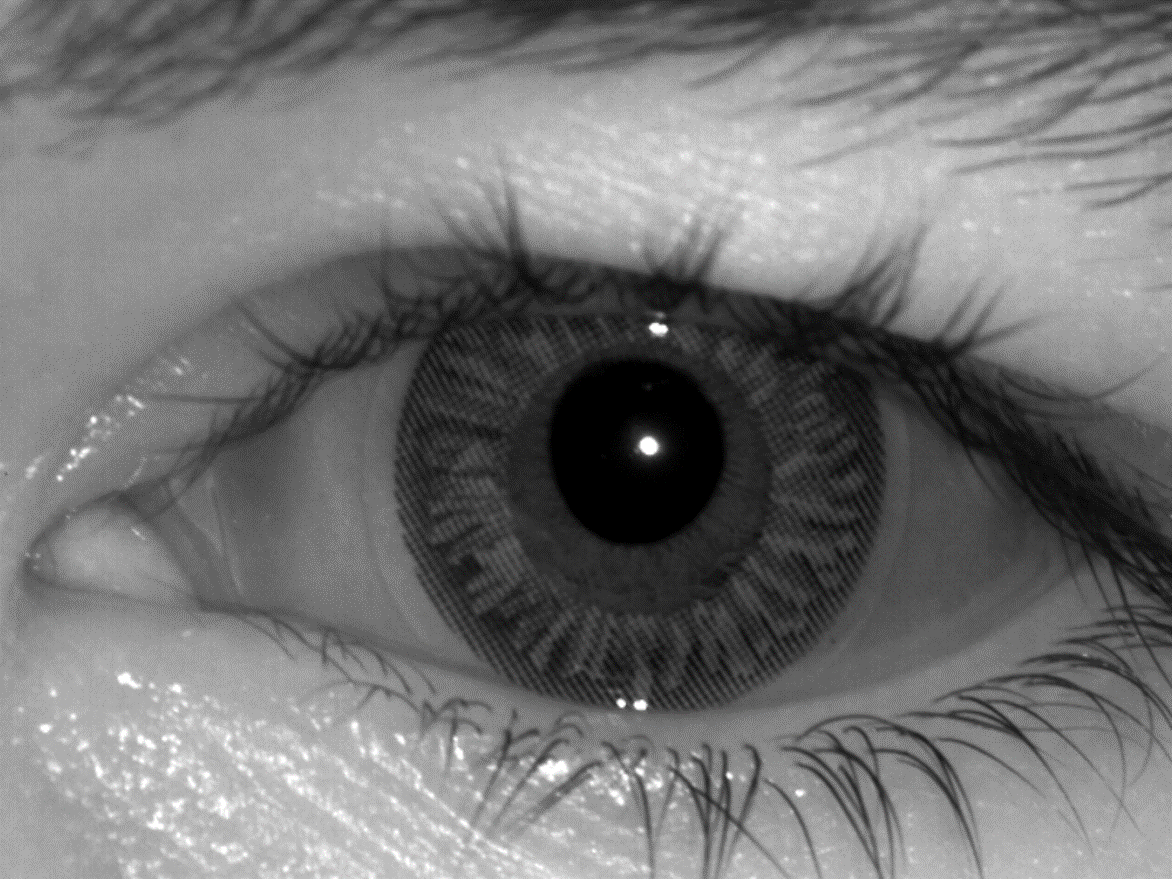}
    \end{minipage} \hspace{0.01\textwidth}
    \begin{minipage}{0.23\textwidth}
        \includegraphics[width=\linewidth]{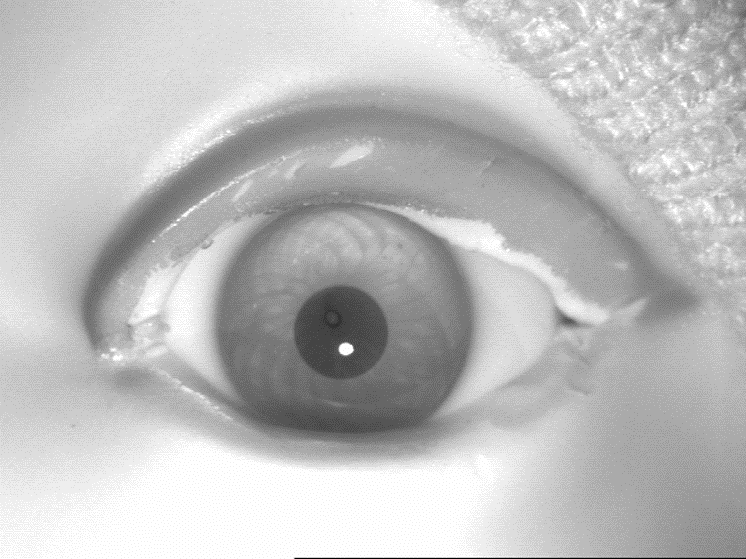}
    \end{minipage}

    \caption{From left to right (with third-party dataset sources, where appropriate): live iris (with no abnormalities) \cite{Kohli_BTAS_2016}, StyleGAN2-generated sample, StyleGAN3-generated sample, iris wearing textured contact lens, then printed and re-captured in near infrared light \cite{Kohli_BTAS_2016}, synthetic sample generated by a non deep learning-based algorithm \cite{CASIA_Synth}, diseased eye \cite{Trokielewicz_BTAS_2015}, glass prosthesis, post-mortem sample \cite{Trokielewicz_TIFS_2019}, iris printout \cite{Czajka_MMAR_2013}, iris wearing textured contact lens \cite{Doyle_ICB_2013}, and artificial eye \cite{Kim_ESA_2016}.}
    \label{fig:attackTypes}
\end{figure*}

\begin{figure*}[!ht]
\centering
\begin{minipage}{0.49\textwidth}
    \centering
    \includegraphics[width=\linewidth]{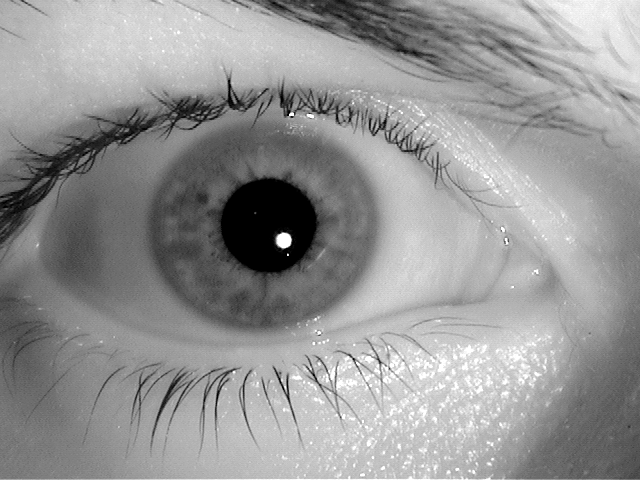}
    \vspace{1.5em}

    \begin{ieeebox}
    {\footnotesize
    \textbf{Examiner Responses:}
    \begin{itemize}[leftmargin=*]
    \item Expert Examiner: ``I think this one is normal. I see specular highlights, the main specular highlight, and then two Purkinje reflections which I think are in the correct place so it will be difficult for StyleGAN to mimic this I think. There's a reflection from the nose on the right, the eyelashes are okay. Final answer normal.''\vskip1mm
    \item Non-Expert Examiner: “Abnormal. Something seems off about that iris. Maybe it's the darker color surrounding the iris, but I don't like that it has, that it looks like they have spikes around it. Other than that, it seems legit. I just really don't like the pupil here. Final answer abnormal.”
    \end{itemize}
    }
    \end{ieeebox}
\end{minipage}
\hfill
\begin{minipage}{0.49\textwidth}
    \centering
    \includegraphics[width=\linewidth]{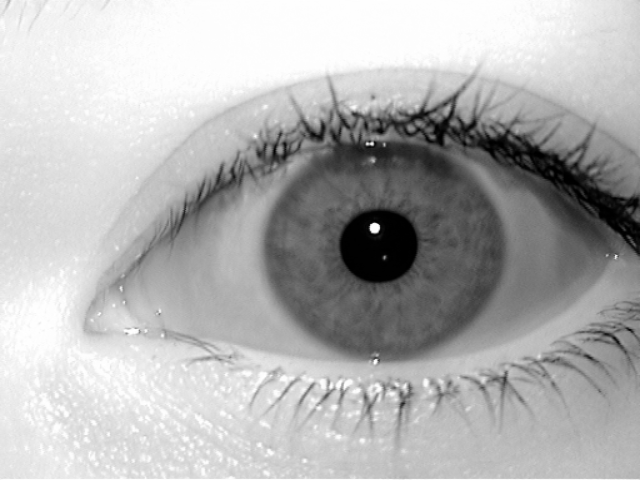}
    \vspace{0.1em}

    \begin{ieeebox}
    {\footnotesize
    \textbf{Examiner Responses:}
    \begin{itemize}[leftmargin=*]
    \item Expert Examiner: ``This one is difficult, but I would say normal. I see the Purkinje reflection which is a good sign. But I also see something weird happening with the eyelids and eyelashes near the corner which means that maybe it is a synthetic image. StyleGAN actually did a really good job here in this case. So I think I will change my mind. Final answer Abnormal.''
    \item Non-Expert Examiner: ``This looks normal to me. Wait, no it doesn't. There's flash in weird parts of the eye on the top and bottom of the iris and on the left and right too, not on the iris but in the corners of the eye. There's also that like dark patch on top, so I think this is abnormal, final answer.''
    \end{itemize}
    }
    \end{ieeebox}
\end{minipage}
\caption{Authentic (left) and synthetically generated by a StyleGAN2 model (right) iris images along with the expert and non-expert descriptions (``Human Text Annotations'' in Fig. \ref{fig:teaser}).}
\end{figure*}

\section{Experimental Design}

\subsection{Experiments}

\subsubsection{Prompts Design}
We performed sixteen experiments pertaining to our research questions. In each experiment we test either Gemini 2.5 Pro model \cite{Google2025_Gemini} (hereafter referred to simply as ``Gemini'') or Llama3.2-Vision:90b model \cite{grattafiori2024llama} (hereafter referred to simply as ``Llama'') in the task of binary iris PAD, once with each of eight prompt variants. During an experiment, for each dataset test image, we send the model a multi-modal prompt consisting of the test image and a text prompt. As part of the prompt in each experiment we explicitly request a floating point number from zero to one (inclusive), with $0.0$ representing a normal iris and $1.0$ representing an attack sample. By thresholding the responses at $0.5$, we obtain a binary classification for each test image, comparable to a traditional CNN-based classifier.

\begin{table}[!htb]
\centering
\caption{Prompt configurations used in experiments with each of the models considered in this work (Gemini and Llama).}
\label{tbl:prompt_overview}
\begin{tabular}{l|c|c}
\toprule
\textbf{Prompt} & \textbf{In-Context} & \textbf{Salience} \\ 
\textbf{Type} & \textbf{Learning} & \textbf{Type} \\ \midrule
Short & & None \\ \hline
Short + Human & \checkmark & Human description  \\ \hline
Short + Llama & \checkmark & Llama MESH  \\ \hline
Short + Gemini & \checkmark & Gemini MESH \\ \hline
Long & & None \\ \midrule
Long + Human & \checkmark & Human description  \\ \hline
Long + Llama & \checkmark & Llama MESH  \\ \hline
Long + Gemini & \checkmark & Gemini MESH \\ \bottomrule
\end{tabular}
\end{table}

Table \ref{tbl:prompt_overview} summarizes eight prompt configurations used in experiments with each of the two models considered in this work (hence sixteenth experiments). Full prompt and salience examples are presented in Supplementary Materials. Each prompt consists of one of two base prompts and either an appended form of salience or no salience. \newtext{When there is no salience the experiment is an example of zero-shot learning. With salience, it is few-shot learning.} The first and most basic prompt is the \textbf{short prompt}, which simply asks the model \textit{``Is this a real and healthy iris or a synthetic/unhealthy iris?''} and then prompts the floating point number. The second base prompt is the \textbf{long prompt}, which was crafted in an expert-in-the-loop feedback session using Claude Sonnet 4 \cite{Anthropic2025_Claude}, with the following structure:

\begin{ieeebox}
\footnotesize
\centering
\textbf{Role and Task:} Take on the role of an expert biometrics examiner and understand the iris PAD task as described.

\vskip2mm
\textbf{Analysis Framework and Classification Context:} Provide an assessment including but not limited to texture, reflections, artifacts, lighting, and other anomalous indicators.

\vskip2mm
\textbf{Required Output Format and Instructions:} Specifically provide a classification label, related confidence, and short explanation of the decision.

\end{ieeebox}

As shown in Table \ref{tbl:prompt_overview}, we also test appending salience to the base prompts, one salience entry for each attack type derived from a sequestered dataset. The ``Human description'' salience refers to raw transcriptions from humans performing iris PAD as described in Sec. \ref{sssec:verbal_human_saliency_acquisition}. The ``Llama MESH'' and ``Gemini MESH'' refer to text descriptions of a given image being classified created by the MLLM as described in Sec. \ref{sssec:MESH_creation}.

\subsubsection{Accessing the MLLMs}
We accessed Gemini through Google's API and Llama through the OpenWebUI API. In both cases, we use the default values for temperature, unlimited maximum tokens (including thinking tokens), and automatically retry until a floating point number is included in the model's response.

\subsubsection{Iris PAD Baselines}
The CNN baseline used in this work is the iris PAD-specific model that uses human salience to guide model training \cite{boyd2023cyborg}. When training our instance of the CNN, we followed the procedure outlined in Boyd's work: 10 independently trained model weight sets (for statistical assessment of the training-related uncertainty), architecture instantiated from DenseNET-121 model \cite{densenet121_2017} (for comparison purposes we do not test other architectures), Stochastic Gradient Descent optimizer with a learning rate of $0.0005$ used, each model trained and tested in a leave-one-out attack type scenario for 50 epochs, and the use of the final weights obtained (after 50 epochs).

We do not replicate the human recognition experiments (to serve as a human recognition baseline), and instead we report (with the original authors' permission) the values from \cite{boyd2022human}. According to these numbers, humans are exceptionally good at solving a binary iris PAD task, making this baseline much stricter than the CNN baseline and provides the target group for our test models to outperform.  

\subsection{Metrics}

Traditional, threshold-based evaluation for iris PAD recommended by ISO/IEC 30107 would include Attack Presentation Classification Error Rate at a \newtext{\textit{chosen}} Bona fide Presentation Classification Error Rate \oldtext{(APCER at BPCER)} \cite{iso30107} \oldtext{\cite{czajka2018presentation}}.
However, we find it necessary to grade APCER and BPCER at a \newtext{\textit{fixed}} threshold of 0.5 for each attack type and live samples and then aggregate these scores for each model using Mean Squared Error. We motivate this choice by the fact that MLLM confidence outputs are typically sparse and discrete, e.g., 0.2, 0.8, 0.95, 1.0 (as illustrated later in Fig. \ref{fig:hist}), \newtext{as they are not a calculation, but a response to the prompt,} making continuous threshold analysis unreliable \newtext{(such as Detection-Error-Trade-off curve)}. For each class the target APCER or BPCER is $0.00$ and the aggregate is thus a measure of how far above that perfect score the model achieves over all classes. The worst possible MSE is thus $1.00$ and the best possible is $0.00$. 


As there is only one set of human responses, one set of Llama-based responses, and one set of Gemini-based responses, we do not calculate a mean or standard deviation for these values. However, for the CNN baseline we report the mean value and standard deviation obtained for 10 independent train-test runs.

\subsection{Datasets}

\subsubsection{Iris Test Images}

\begin{figure}[!ht]
    \centering
    \includegraphics[width=0.6\linewidth]{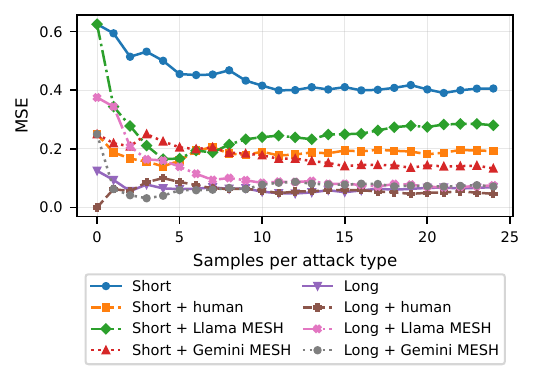}
    \caption{Learning curve analysis of the eight Gemini experiments showing the MSE scores converge before even 25 samples from each attack type.}
    \label{fig:learning_curve}
\end{figure}

\oldtext{In order to make a valid comparison with the human and CNN baselines, we use the established iris PAD test set with human-annotated saliency maps from \cite{boyd2023cyborg} consisting of the following classes: normal iris, artificial eyes (\eg glass prosthesis), iris printouts, eyes wearing textured contacts, printouts of irises wearing textured contact lenses (later referred simply to ``contacts+print''), diseased eyes, post-mortem eyes, and synthetically-generated iris images  \cite{real_iris1}, \cite{real_artificial_textured_print_24}, \cite{real_print12}, \cite{real_diseased}, \cite{real22}, \cite{real_textured20}, \cite{real_textured46}, \cite{real43}, \cite{real_textured45}, \cite{post40}, \cite{print21}, \cite{synth44}, \cite{diseased38}.} 

\newtext{In order to make a valid comparison with the human and CNN baselines, we use the established iris PAD test set with human-annotated saliency maps from \cite{boyd2023cyborg} consisting of the following classes: (a) normal iris \cite{real_iris1,real_artificial_textured_print_24,real_print12,real_diseased}, \cite{real22,real_textured20,real_textured46,real43}, \cite{real_textured45}, (b) artificial eyes (\eg glass prosthesis) \cite{real_artificial_textured_print_24}, (c) iris printouts \cite{real_artificial_textured_print_24,real_print12,print21}, (d) eyes wearing textured contacts \cite{real_artificial_textured_print_24,real_textured20,real_textured46,real_textured45}, (e) printouts of irises wearing textured contact lenses (later referred simply to ``contacts+print'') \cite{real_artificial_textured_print_24}, (f) diseased eyes \cite{real_diseased,diseased38}, (g) post-mortem eyes \cite{post40}, and (h) synthetically-generated iris images \cite{synth44}.}

However, due to cost constraints associated with running commercial MLLM (Gemini) we sample 30 images from each class (or all images, if there are less than 30) to make a final dataset of 224 test images \newtext{(See Tab. \ref{tab:attack-types})}. We confirmed the statistical validity of our test sample size in two ways: 1) we performed a learning curve analysis (see Fig. \ref{fig:learning_curve}) wherein MSE is repeatedly recalculated as samples are added to see that the metric converges, and 2) by using the Wilcoxon \newtext{\cite{Conover1998}} and Mann-Whitney-U \newtext{\cite{Fay2010}} tests to compare multiple runs of samples (see Supplementary Materials for the full statistical analysis). \newtext{We evaluate the models with a low number of samples, because we leverage the abilities of MLLMs trained on massive amounts of data. The purpose of this evaluation is to estimate generalization performance on such a set-up. We further recommend this kind of analysis for works with access to small sample sizes.} 

Note that the iris PAD test set is image-disjoint from \newtext{the provided training set in \cite{boyd2023cyborg} used to recreate the CNN baseline and} the set used to derive the human and machine-expanded saliency from human (MESH) salience, described in the following subsections.

\begin{table}[t]
\centering
\caption{\newtext{Attack types and number of samples.}}
\label{tab:attack-types}
\begin{tabular}{l r}
\hline
\newtext{Attack type} & \newtext{Number of samples} \\
\hline
\newtext{Live iris} & \newtext{30} \\
\newtext{Artificial} & \newtext{14} \\
\newtext{Contacts+print} & \newtext{30} \\
\newtext{Diseased} & \newtext{30} \\
\newtext{Post-mortem} & \newtext{30} \\
\newtext{Printout} & \newtext{30} \\
\newtext{Synthetic} & \newtext{30} \\
\newtext{Textured contact} & \newtext{30} \\
\hline
\end{tabular}
\end{table}

\subsubsection{Verbal Human Saliency Acquisition}
\label{sssec:verbal_human_saliency_acquisition}

We conducted a data collection consisting of 70 non-experts and 8 experts in iris PAD.
Participants were tasked with identifying samples as either ``normal'' or ``abnormal,'' and then were asked to verbally describe what led them to their decision. \newtext{To ensure coherency, participants were given definitions for normal and abnormal that matched the division between attack (artificial, printout, diseased, post-mortem, synthetic, and irises with textured contacts) and non-attack (live and healthy irises) samples.} To record the verbal descriptions participants wore a MeetSummer X1 Lavalier wireless microphone \newtext{(purchased in February 2024)}. 
Ethnicity, gender, and age were documented for all participants. Expert participants were recruited from university faculty and PhD students doing active research in iris recognition or iris PAD for at least one year, while the non-expert participants consisted of university students and faculty who did not possess any biometric research experience. The study was approved by the Institutional Review Board (IRB), and each participant signed a consent form allowing us to release anonymized data. 

Iris images used in the human saliency collection were sourced from the dataset used previously in a study conducted by Boyd \etal \cite{boyd2022human}. In addition to these images, 100 synthetic iris samples were generated using StyleGAN2 and StyleGAN3 with 50 for each category \cite{karras2020analyzing, karras2021alias, tinsley2023iris}. These images were divided up into 16 unique decks that were carefully balanced to contain the same number of samples per anomaly type ensuring that all decks were equal in difficulty and to ensure sufficient collection of verbal descriptions for each sample.

Each session began with a brief training, regardless of expertise, to ensure everyone had a baseline understanding of what constitutes normal and attack samples. Included during this training were sample verbal descriptions. The samples shown in the training process were excluded from the actual experiment. Once the session started, each participant was shown a single image per slide and was instructed to first inspect the image and provide a classification of either normal or attack. Next, the participants were instructed to verbally justify their answers. The instructions for the verbal description phase were intentionally open-ended as not to instill any arbitrary guidelines that may skew the authenticity of the results. Participants were advised to be as thorough as they felt necessary and that there was not a time limit imposed. Instead they were encouraged to focus on what features stood out during the initial phase or what their thought process was when diagnosing the image. Upon completion of the verbal description phase, participants announced their final answer followed by the ``done'', at which point the moderator advanced to the next sample until completion.

\subsubsection{Data Post-Processing}
The audio recordings of each participant's session were transcribed to text using a local installation of Whisper \cite{radford2023robust} and manually inspected to correct any transcription errors while maintaining verbal filler words to maintain the authenticity of the original descriptions. These final corrected prompts received the designation of human saliency descriptions and are verbal saliency source for the human-injected experimental results.

The next step was to create the Machine-Expanded Saliency from Human (MESH) descriptions. The inspiration for MESH descriptions came from observations in the way the majority of participants approached the verbal description phase. Our guidelines for the verbal description phase were for participants to identify what features stood out as being critical for their decision making or their thought process during the evaluation. We did not want to impose strict guidelines in an effort to avoid elevating features that were not critical to the decision making process for the sample. While this decision preserves only the most important features and observations, it leaves the verbal descriptions incomplete when viewed as an \textit{image description}. When reading an individual human saliency description without the iris sample in view, the context is missing. However, MESH descriptions are intended to be a thorough and comprehensive image description that is elevated by the observations of the human subjects. As a result, we leveraged the help of two MLLMs, Llama3.2-Vision:90b and Gemini 2.5 Pro \cite{grattafiori2024llama, Google2025_Gemini}, to help turn the human saliency descriptions into a robust image description that when read independent of the image, are much more thorough in explaining the image as a whole. The end result is two different sets of MESH descriptions for each image in the dataset referred to as ``Llama MESH'' and ``Gemini MESH''. Due to the sensitive nature of biometric data, our choice of MLLM models to generate the MESH descriptions were limited to models either approved by our institution's compliance team, or models deployed within a local environment.

\subsubsection{MESH Prompt Creation}
\label{sssec:MESH_creation}
To make the MESH prompts, we fine-tuned a prompt with the assistance of Claude Sonnet 4 \cite{Anthropic2025_Claude}. After explaining the objective to Claude and sharing iris samples and paired descriptions not constrained by the IRB, we tested variations of the prompts on Llama until we were satisfied with the results. Claude was given the task of designing a prompt that would be given to Llama paired with an image to extract a MESH description for that sample. Claude was given an eye sample not constrained by any existing data licensing agreement or IRB, examples of human examiner observations, and a final comprehensive image description to demonstrate the desired output from the prompt it was tasked with creating. Upon creation, the prompt was tested on Llama with relevant samples and this feedback was relayed to Claude explaining what needed to be corrected in the form of revising the prompt structure to obtain the desired outcome. 
Examples of revisions included diseased irises initially being labeled as normal instead of attack, agreement bias where the model would simply agree with what an expert said regardless of how illogical the observations were, and not properly synthesizing all observations into a seamless final description. The final prompt structure was as follows:

\begin{ieeebox}
\footnotesize
\centering
\textbf{Analysis Framework:} Provide an initial assessment by examining things including but not limited to texture, reflections, artifacts, lighting, and other spoofing indicators.

\vskip2mm
\textbf{Examiner Feedback Evaluation:} You will receive feedback formatted as examiner\_ID, expertise status, correct or incorrect classification, and verbal description.

\vskip2mm
\textbf{Critical Synthesis:} Validate human examiner observations, weigh expert technical knowledge with higher priority but do consider non-expert intuitive insights, use classification accuracy to calibrate examiner reliability, resolve conflicts by prioritizing verifiable observations.

\vskip2mm
\textbf{Required Output Format:} Image Classification, Confidence, Key Features Observed, Spoofing Indicators, Examiner Integration, Technical Details, Comprehensive Iris Description.
\end{ieeebox}

At a high level, the MESH description process starts with submitting the image to the model, which assesses the image according to the analysis framework and generates a baseline description of the sample. Next, the model, either Llama or Gemini, considers human examiner feedback and looks for agreement between its own observations and the human's observations with an emphasis towards correct human observations and priority to expert observations. Once complete, the analysis is revised into a final comprehensive description that is a thorough description of the image in its entirety, informed from human examiners, as opposed to specific feature observations. This process was performed using both Llama and Gemini to test for any performance differences between the two MLLMs, particularly as the models differ in context size, architecture, and proprietary status.

\section{Results}

\begin{table}
    \centering
    \caption{MSE scores for Gemini and Llama based on the prompt type used. The best prompt variant of the short and long prompts is bolded for each model.}
    \begin{tabular}{c c c}
        \toprule
        \textbf{Model} & \textbf{Prompt variant} & \textbf{MSE}  \\
        \midrule
        Human Subjects & ~                       & 0.062              \\
        \midrule
        Baseline CNN   & ~                       & $0.345 \pm 0.041$  \\
        \midrule
        ~              & Short                   & 0.416              \\
        Gemini         & Short + Human           & 0.183              \\
        ~              & Short + Llama           & 0.273              \\
        ~              & Short + Gemini          & {\bf 0.118}        \\
        \midrule
        
        ~              & Long                    & 0.062              \\
        Gemini         & Long + Human            & {\bf 0.053}        \\
        ~              & Long + Llama            & 0.087              \\
        ~              & Long + Gemini           & 0.078              \\
        \midrule
        ~              & Short                   & 0.422              \\
        Llama          & Short + Human           & 0.190              \\
        ~              & Short + Llama           & 0.267              \\
        ~              & Short + Gemini          & {\bf 0.125}        \\
        \midrule
        ~              & Long                    & 0.411              \\
        Llama          & Long + Human            & {\bf 0.074}        \\
        ~              & Long + Llama            & 0.098              \\
        ~              & Long + Gemini           & 0.125              \\
        \bottomrule
    \end{tabular}
    \label{tab:MSE_table}
\end{table}

\begin{sidewaystable}
\scriptsize
\centering
\caption{BPCER and APCER scores at threshold 0.5 for live iris and each iris PAD attack type by \textit{Model} (Gemini or Llama) and \textit{Prompt variant}. Columns correspond to the attack types.}
\begin{tabular}{c c c c c c c c c c}
\toprule
\textbf{Model} & \textbf{Prompt variant} & \textbf{Live iris} & \textbf{Artificial} & \textbf{Contacts+print} & \textbf{Diseased} & \textbf{Post-mortem} & \textbf{Printout} & \textbf{Synthetic} & \textbf{Textured contact} \\
\midrule
Human Subjects & ~ & 0.437 & 0.349 & 0.057 & 0.020 & 0.162 & 0.298 & 0.118 & 0.227 \\
\midrule
Baseline CNN & ~ & $0.000\pm0.000$ & $0.236\pm0.143$ & $0.990\pm0.016$ & $0.227\pm0.099$ & $0.900\pm0.061$ & $0.647\pm0.118$ & $0.377\pm0.146$ & $0.460\pm0.168$ \\
\midrule
~ & Short & {\bf 0.000} & 0.769 & 0.700 & 0.833 & 0.833 & 0.267 & 0.833 & 0.300 \\
Gemini & Short + Human & 0.033 & {\bf 0.077} & 0.567 & {\bf 0.700} & 0.400 & 0.167 & 0.633 & {\bf 0.233} \\
~ & Short + Llama & 0.033 & 0.462 & 0.667 & {\bf 0.700} & 0.700 & 0.167 & 0.667 & 0.267 \\
~ & Short + Gemini & 0.200 & {\bf 0.077} & {\bf 0.333} & 0.767 & {\bf 0.367} & {\bf 0.133} & {\bf 0.067} & 0.200 \\
\midrule
~ & Long & 0.167 & 0.083 & {\bf 0.222} & 0.517 & 0.233 & {\bf 0.074} & 0.241 & 0.172 \\
Gemini & Long + Human & 0.300 & {\bf 0.000} & 0.233 & {\bf 0.433} & {\bf 0.133} & 0.133 & {\bf 0.167} & {\bf 0.167} \\
~ & Long + Llama & {\bf 0.133} & 0.077 & 0.400 & 0.533 & 0.300 & 0.167 & 0.267 & 0.200 \\
~ & Long + Gemini & 0.367 & {\bf 0.000} & 0.233 & 0.467 & 0.300 & 0.100 & 0.300 & {\bf 0.167} \\
\midrule
~ & Short & {\bf 0.167} & 0.692 & 0.759 & 0.800 & 0.533 & 0.467 & 0.724 & 0.793 \\
Llama & Short + Human & 0.233 & 0.385 & 0.633 & 0.433 & 0.400 & 0.433 & 0.400 & 0.467 \\
~ & Short + Llama & 0.400 & 0.583 & 0.643 & 0.533 & 0.433 & 0.433 & 0.400 & 0.633 \\
~ & Short + Gemini & 1.000 & {\bf 0.000} & {\bf 0.000} & {\bf 0.000} & {\bf 0.000} & {\bf 0.000} & {\bf 0.000} & {\bf 0.000} \\
\midrule
~ & Long ~ & {\bf 0.048} & 1.000 & 0.688 & 0.667 & 0.412 & 0.381 & 0.529 & 0.880 \\
Llama & Long + Human & 0.640 & 0.286 & 0.115 & 0.208 & 0.048 & {\bf 0.000} & 0.050 & 0.200 \\
~ & Long + Llama & 0.500 & 0.429 & 0.263 & 0.364 & 0.105 & 0.053 & 0.143 & 0.333 \\
~ & Long + Gemini & 1.000 & {\bf 0.000} & {\bf 0.000} & {\bf 0.000} & {\bf 0.000} & {\bf 0.000} & {\bf 0.000} & {\bf 0.000} \\
\bottomrule
\end{tabular}
\label{tab:recall_table}
\end{sidewaystable}

\subsection{\textbf{Answering RQ1:} \printRQaShort}

As seen in Tab. \ref{tab:MSE_table}, given only the short prompt with no guiding salience, Gemini begins to approach, but does not surpass the traditional CNN baseline with a MSE of $0.416$ over $0.345$. When viewing Gemini's per-attack-type results with the short prompt variant, we see that it is able to detect some attack types innately. It correctly identifies all of the live irises and performs better than the human baseline on printout attacks. It outperforms the CNN baseline on 4 of 8 attack types: contacts+print, post-mortem, printouts, and textured contacts. Llama performs comparably to Gemini in terms of the MSE.

Hence, the answer to \textbf{RQ1} is \textbf{mixed}. While Gemini and Llama are able to outperform the human and CNN baselines for some attack types, it is overall worse than both.

\subsection{\textbf{Answering RQ2:} \printRQb}

Using the long, engineered prompt to test the MLLMs yields much better results for Gemini, but almost no improvement for Llama. Gemini both surpasses the traditional CNN and matches the human results with a MSE of $0.062$ against scores of $0.345$ and $0.62$ respectively. Overall, Gemini with the long prompt is the second-best model tested.

On a per-attack-type basis Llama outperforms the human baseline and Gemini with the long prompt only on identifying normal irises, but outperforms the CNN on two attack types: contacts+print and post-mortem. Gemini outperforms the human subjects on normal, artificial, printout, and textured contacts (half of the categories). It outperforms the CNN on all but normal and diseased irises.

Hence, the answer to \textbf{RQ2} is that a structured prompt \textbf{can improve general MLLM performance in the task of iris PAD to exceed that of comparable CNNs and even human subjects}.

\subsection{\textbf{Answering RQ3:} \printRQcShort}

In all cases, including human salience examples in the prompt improves performance over the base prompt. Gemini using the short prompt with human salience outperforms the baseline CNN with a MSE value of $0.183$ against $0.345$. Gemini using the long prompt with human salience performs the best of all tests and is the only experiment to outperform the human subjects with a MSE value of $0.053$ against $0.062$. Llama also sees a performance boost, outscoring the baseline CNN with both short and long prompts with human salience.

On a per-attack-type basis, Llama with a long prompt + human salience beats the baseline CNN on five of eight categories. Gemini with a long prompt + human salience outperforms the human subjects on all but three attack types: contacts+print, diseased, and synthetic.

Thus, the answer to \textbf{RQ3} is \textbf{just as human salience allowed for CNNs to improve in the task of iris PAD, injecting human salience into MLLM prompts allows both models to outperform the comparable CNN and Gemini to outperform even humans.}

\subsection{\textbf{Answering RQ4}: \printRQdShort}

In all cases, including a MESH (either from Llama or Gemini) improves MLLM performance to be better than the baseline CNN. However, neither MLLM improves over the human baseline with the MESH injected prompts. Only in the case of using the short prompt does the inclusion of MESH improve over using prompt + human salience. This higher performing MESH came from Gemini, the Llama MESH resulted in a lower performance boost.

\subsection{Observations}

\begin{figure}[!ht]
  \centering
  \begin{subfigure}[t]{\linewidth}
    \centering
    \includegraphics[width=0.6\textwidth]{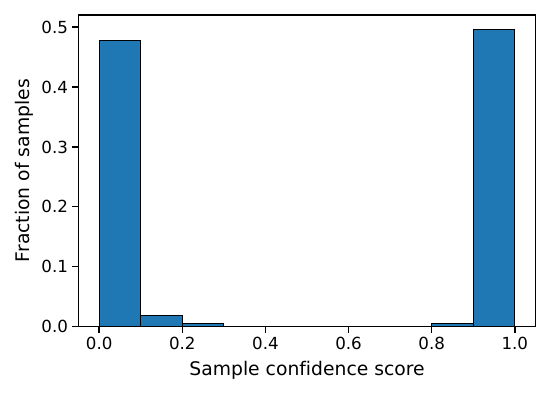}
    \caption{Gemini using short prompt with human salience.}
    \label{fig:hist:a}
  \end{subfigure}\\
  \begin{subfigure}[t]{\linewidth}
    \centering
    \includegraphics[width=0.6\textwidth]{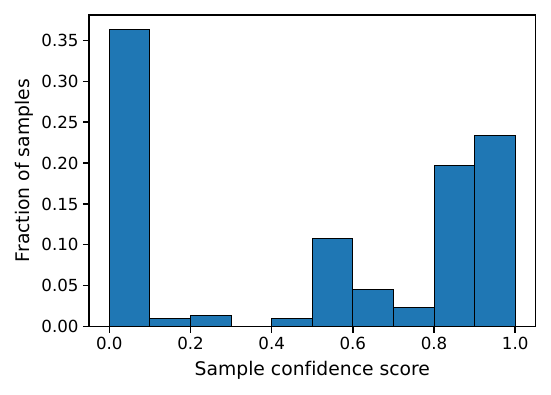}
    \caption{Llama using short prompt with human salience.}
    \label{fig:hist:b}
  \end{subfigure}
  \caption{Histograms of the test set iris PAD classification confidence values generated by the two MLLMs. We see that Llama is more capable of expressing uncertainty.}
  \label{fig:hist}
\end{figure}

We note several key observations from all of our experiments:

\vskip1mm\noindent\textbf{a) Including human salience in well-engineered prompts achieves state-of-the-art iris PAD performance.}
Including human exemplars not only outperforms traditional CNNs trained with human visual salience, but can allow for larger models to even outperform humans.

\vskip1mm\noindent\textbf{b) In absence of a well-engineered prompt, MLLMs can expand on salience to their own benefit.} The short prompt represents a poorly engineered prompt. In this case, while including human salience did improve performance (Table \ref{tab:MSE_table}), it did not improve performance as well as MESH salience. The MLLMs are able to expand on human descriptions with task-relevant details for helping themselves.

\vskip1mm\noindent\textbf{c) Gemini performs better but local Llama is more capable of expressing uncertainty.} While Gemini's best performance is ultimately a lower MSE than Llama's, neither are perfect. In Fig. \ref{fig:hist} we see that Gemini produces essentially bimodal confidence values, while Llama is more able to place it's confidence in the middle of the range. (Histograms for all experiments are available in Supplementary Materials.) 

\vskip1mm\noindent\textbf{d) General MLLMs are capable of overfitting.} For \oldtext{Gemini}\newtext{Llama}, both the short and long prompts with Gemini MESH salience led to classifying every iris as an attack attempt. \newtext{The Gemini MESH is the longest prompt and this overfitting may be due to the length. The Gemini Mesh contains on average three times more text tokens than that of the Llama MESH (estimates made using OpenAI tokenizer \cite{OpenAI_Tokenizer}, and both of these saliency types add significantly more tokens to the overall text input. However, context window size is not an issue as we estimate the maximum prompt, including the image, as only 16,000 tokens, well below Llama’s 128,000 token limit. Recent findings from Sun \emph{et al.} \cite{sun2025mitigating} suggest that MLLMs suffer visual attention degradation as long-chain text reasoning increases. While the models tested in Sun's experiments did not include Llama 3.2-vision:90b, many of the models tested that suffer from this problem are all CLIP-based \cite{radford2021learning} vision backbones and share similar architectures to that of Llama according to their PyTorch implementations \cite{paszke2019pytorch,torchtune}.}

\section{Limitations and Future Work}

The principle complications of this study are related to (a) expense and (b) compliance.
Due to the sensitive nature of biometric (iris in this case) data we used the only commercial model for which we could obtain a security guarantee, based on institutional bilateral agreement. Such agreements are rare, and thus the portfolio of commercial MLLM models possible to be used without compromising subjects' privacy and thus breaking the IRB protocols, is very small.
Due to the expense of running Gemini and hosting Llama we sub-sampled our dataset to the smallest, but still statistically viable amount.

Furthermore, this study only exhibits the two extremes of prompting: the short, minimal prompt, and the long, expert-engineered prompt. Future work will delve into the components of prompting to bridge the gap between these two.

As Llama exhibited the ability to express uncertainty, other future works will explore MLLM calibration and adversarial robustness.
We also note that model strengths tend to be complementary across the different attack types, indicating the potential for future ensemble systems.
Last but not least, a further study should explore the effect of using MLLMs to justify decisions to human examiners as part of trustworthy AI.

\section{Conclusion}


The main, and important conclusion from this work is that generalist MLLMs exhibit the ability to perform a rather niche iris spoofing detection task with a scaffold of well-structured prompts augmented with human salience.
These MLLMs can consistently outperform salience-based CNNs and the best MLLM prompt variant, Gemini with the long, well-engineered prompt with raw human textual information, can outperform human subjects.

Prompt engineering makes a significant difference for this specialized task. 
The short prompt variant performed poorly regardless of MLLM, however, with the addition of MESH salience, it could achieve a lower MSE than the CNN.

Ultimately, prompting with human salience makes using an off-the-shelf MLLM, even a local one, viable for iris PAD and opens up opportunities for new implementations in the field.

This paper offers a dataset of human verbal descriptions for the existing iris PAD benchmark \cite{boyd2022human} to facilitate follow-up future studies and replicability of this work\footnote{Instructions on how to request a copy of the dataset will be available at \url{https://github.com/CVRL/Multimodal-LLMs-Biometric-Expertise} in the event of this paper being accepted}.

\section{Acknowledgments}
This work was supported by the U.S. Department of Defense (Contract No. W52P1J-20-9-3009) and by the National Science Foundation (Grant No. 2237880). Any opinions, findings, and conclusions or recommendations expressed in this material are those of the authors and do not necessarily reflect the views of the National Science Foundation, the U.S. Department of Defense or the U.S. Government. The U.S. Government is authorized to reproduce and distribute reprints for Government purposes, notwithstanding any copyright notation here on.

\printbibliography

\end{document}